\documentclass[journal]{IEEEtran}
\usepackage[utf8]{inputenc}
\usepackage{cite}
\usepackage{hyperref}
\usepackage{multirow}
\usepackage{algorithmic}
\usepackage{orcidlink}
\usepackage{subfigure}
\usepackage{amsmath} 
\usepackage{amsfonts,amssymb}  
\usepackage{colortbl}
\graphicspath{ {./figures/} }
\begin{document}

\title{Sparse Deformable Mamba for Hyperspectral Image Classification}

\author{Lincoln Linlin Xu, ~\IEEEmembership{Member,~IEEE}, Yimin Zhu, Zack Dewis, Zhengsen Xu, Motasem Alkayid, Mabel Heffring, Saeid Taleghanidoozdoozan 


\thanks{This work was supported by the Natural
Sciences and Engineering Research Council of Canada (NSERC) under Grant RGPIN-2019-06744.}
\thanks{Lincoln Linlin Xu, Yimin Zhu, Zack Dewisis, Zhengsen Xu, Mabel Heffring, Saeid Taleghanidoozdoozan, are all with the Department of Geomatics Engineering, University of Calgary, Canada (email: (lincoln.xu, yimin.zhu, zachary.dewis, zhengsen.xu, mabel.heffring1)@ucalgary.ca, staleghanidoozdoozan@uwaterloo.ca) (Corresponding author: Lincoln Linlin Xu)}.

\thanks{Motasem Alkayid is with the Department of Geomatics Engineering, University of Calgary, Canada, and also with the Department of Geography, Faculty of Arts, The University of Jordan, Amman, Jordan (email: motasem.alkayid@ucalgary.ca)
}}

\markboth{Journal of \LaTeX\ Class Files,~Vol.~13, No.~9, September~2014}
{Shell \MakeLowercase{\textit{et al.}}: }
\maketitle

\begin{abstract}
Although Mamba models significantly improve hyperspectral image (HSI) classification, one critical challenge is the difficulty in building the sequence of Mamba tokens efficiently. This paper presents a Sparse Deformable Mamba (SDMamba) approach for enhanced HSI classification, with the following contributions. First, to enhance Mamba sequence, an efficient Sparse Deformable Sequencing (SDS) approach is designed to adaptively learn the ''optimal" sequence, leading to sparse and deformable Mamba sequence with increased detail preservation and decreased computations. Second, to boost spatial-spectral feature learning, based on SDS, a Sparse Deformable Spatial Mamba Module (SDSpaM) and a Sparse Deformable Spectral Mamba Module (SDSpeM) are designed for tailored modeling of the spatial information spectral information. Last, to improve the fusion of SDSpaM and SDSpeM, an attention based feature fusion approach is designed to integrate the outputs of the SDSpaM and SDSpeM. The proposed method is tested on several benchmark datasets with many state-of-the-art approaches, demonstrating that the proposed approach can achieve higher accuracy with less computation, and better detail small-class preservation capability.  

\end{abstract}


\IEEEpeerreviewmaketitle

\section{Introduction}

\begin{figure}
    \centering
    \includegraphics[width=0.49\textwidth]{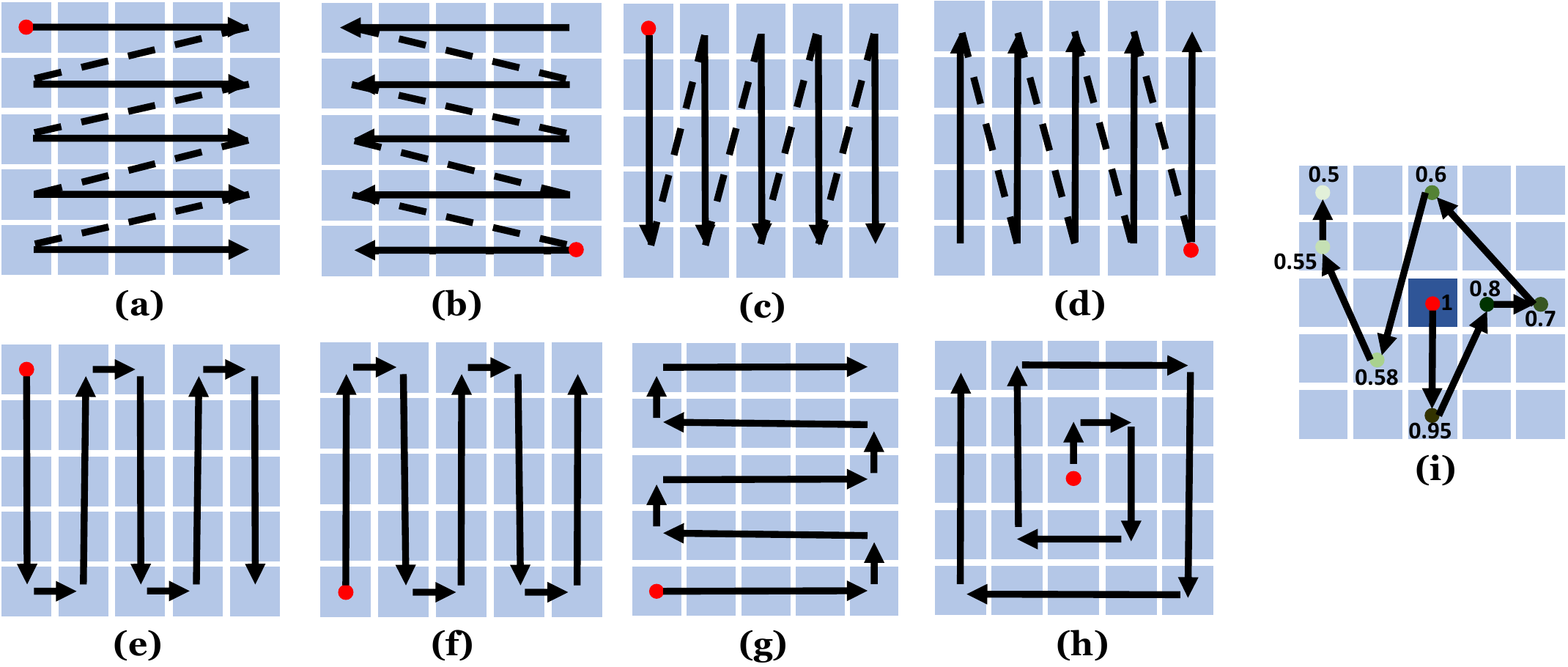}
    \caption{Illustration of our proposed \textbf{sparse deformable sequencing (SDS)} for improving Mamba. Various classical scanning approaches in (a)-(h) are \textbf{dense and predefined}, because they use all tokens in a \textbf{deterministic} manner, causing potential \textbf{redundancy and rigidity, unnecessary computation cost}, and \textbf{difficulty} in selecting scanning approaches. In contrast, our SDS approach in (i) can \textbf{identify and sequence} \textbf{limited number of relevant tokens in a learnable and adaptive manner}, leading to sparse and deformable sequence patterns that can reduce redundancy, rigidity and computational cost in Mamba. }
    \label{fig:sequence}
\end{figure}
Hyperspectral image (HSI) classification is a fundamental task, which transforms raw HSI data into valuable maps that support various key environmental and resource exploitation tasks. Nevertheless, efficient HSI classification is challenging due to various difficult HSI characteristics, e.g., high-dimensionality, noise, spatial-spectral heterogeneity, and limited training samples. Given these difficulties, it is challenging to extract discriminative features that can efficiently capture subtle differences among HSI classes. Therefore, designing efficient feature learning techniques using advanced machine learning (ML) and deep learning (DL) approaches is a critical research topic. 

In previous decades, various approaches have been proposed for feature learning from HSI. For example, principal component analysis (PCA) \cite{rodarmel2002principal} and independent component analysis (ICA) \cite{stone2002independent} have been used to extract compact spectral features from HSI. A morphological approach \cite{benediktsson2005classification} has also been used to enhance the spatial feature extraction. However, these approaches are feature engineering approaches that cannot fully capture the discriminative HSI information in an adaptive manner. Deep learning approaches have been widely used for improving HSI feature extraction. Convolutional neural networks (CNN) \cite{li2021survey, 8061020, 10144309} improve the learning of HSI spatial-spectral features, but they have limitations in terms of strong inductive bias and locality to capture the long-range spatial-spectral correlation effect in HSI. Transformers are more flexible and adaptive to longer-range spatial context \cite{9565208, 10472541, SSFTT}, but they require a large attention matrix and thereby big computational cost. Comparing with Transformers, the Mamba models, due to their token sequencing, can greatly reduce computations while maintaining the long-range modeling capacity. Hence, many Mamba approaches \cite{10604894, 10703171, liu2024hypermamba} have been proposed for improving HSI classification. 

However, for Mamba models \cite{gu2023mamba}, one critical issue is how to build the sequence in a compact and efficient manner. First, compactness is essential for enabling longer-sequence learning and reducing computational cost and redundancy in tokens. Second, the order of tokens in Mamba sequence is critical for improving Mamba's modeling capacity and overcoming the correlation vanishing issue. Therefore, instead of scanning tokens in a dense, predefined, and deterministic manner, defining token sequence in a sparse, adaptive and learnable manner is critical for boosting Mamba modeling capacity.

This paper therefore proposes a sparse and deformable Mamba (SDMamba) model to improve HSI classification, with the following contributions. 
\begin{itemize}

\item To improve compactness and efficiency in Mamba token sequencing, instead of using deterministic dense scanning approaches, a sparse deformable sequencing (SDS) approach is designed, which can identify and sequence a limited number of relevant tokens in a learnable and adaptive manner, leading to sparse and deformable sequence patterns that can reduce redundancy, rigidity, and computational cost in Mamba. As illustrated in Figure \ref{fig:sequence}, compared to various predefined scanning approaches, the proposed SDS approach tends to overcome the potential redundancy and rigidity, unnecessary computational cost, and the difficulty selecting among various scanning approaches.

\item To improve spatial-spectral feature learning, based on SDS, a Sparse Deformable Spatial Mamba Module (SDSpaM) and a Sparse Deformable Spectral Mamba Module (SDSpeM) are designed for tailored modeling of the spatial information spectral information respectively. 

\item To enhance the fusion of SDSpaM and SDSpeM, an attention based feature fusion approach is designed to integrate the outputs of the SDSpaM and SDSpeM. The proposed method is tested several benchmark datasets with many state-of-the-art approaches, demonstrating that the proposed approach can achieve higher accuracy and better detail small-class preservation capability. 

\end{itemize}


The reminder of the paper is organized as follows. Section II illustrates the details of the proposed SDSMamba approach. Section III presents the experimental design and results. Section IV concludes this study.

\section{Methodology}
\label{methodology}
\subsection{Overview of Sparse Deformable Mamba (SDMamba) Model}
\begin{figure*} 
    \centering
    \includegraphics[width=1.03\textwidth]{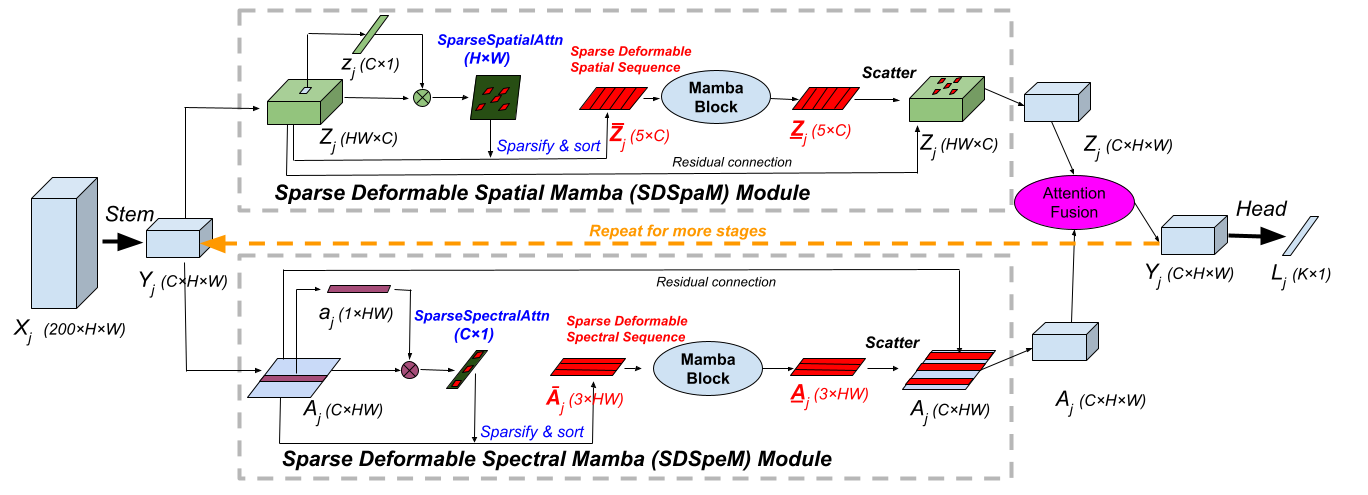} 
    \caption{\textbf{The proposed SDMamab is sparse}, because the input sequences to the \textbf{MambaBlock}, i.e.,  $\overline{\boldsymbol{\mathit{Z}}}_j (5 \ tokens)$ and $\overline{\boldsymbol{\mathit{A}}}_j (3 \ tokens)$ \textbf{have much less number of tokens} than  $\boldsymbol{\mathit{Z}}_j (HW \ tokens)$ and $\boldsymbol{\mathit{A}}_j (C \ tokens)$ respectively. Moreover, in SDMamba, \textbf{the order of tokens in $\overline{\boldsymbol{\mathit{Z}}}_j$ and $\overline{\boldsymbol{\mathit{A}}}_j$ are deformable and learnable}, because two adaptive attention matrices, i.e., $SparseSpatialAttn$ and $SparseSpectralAttn$, are used to sort the tokens and identify a limited number of relevant tokens. Therefore, the proposed SDMamba approach has \textbf{sparse and deformable sequence patterns that can reduce redundancy, rigidity, and computational cost in classical Mamba models}. }
    \label{fig:overview}
\end{figure*}

Figure \ref{fig:overview} displays the architecture of the proposed SDMamba model. As we can see, the proposed SDMamba model ingests $\mathit{X_j}$, a $H \times W \times 200$ data cube, with $W$, $H$ and $200$ being respectively the width, height and number of channels. Taking Indian Pines dataset as an example, $H=9$, $W=9$, and there are 200 channels. After the SDSpaM and the SDSpeM modules and the attention fusion module, it outputs the label of the center pixel of this data cube, i.e., $\mathit{L_j}$, a $K \times 1$ tensor, where $K$ is the number of classes. The key building blocks are sparse deformable sequencing (SDS), SDSpaM, SDSpeM and attention fusion module, which are illustrated in the subsequent subsections.

\begin{figure}
    \centering
    \includegraphics[width=0.4\textwidth]{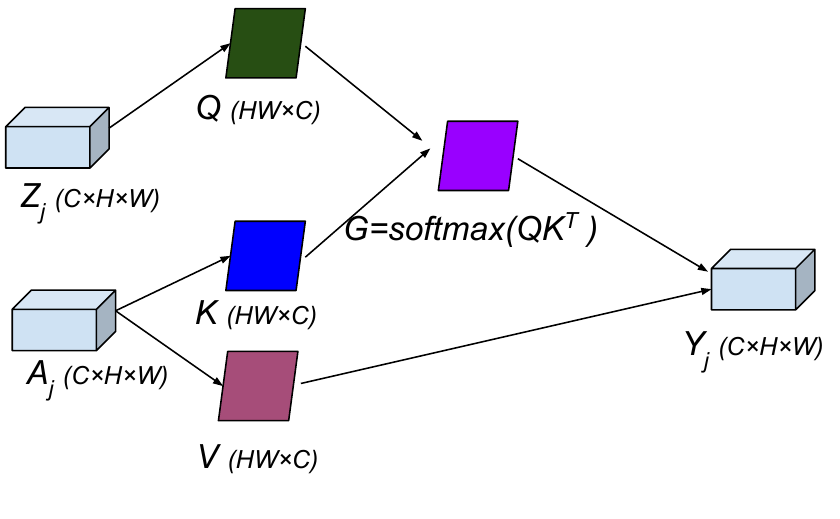}
    \caption{The outputs of SDSpaM and SDSpeM, i.e., $\boldsymbol{\mathit{Z}}_j$, $\boldsymbol{\mathit{A}}_j$ respectively, are fused using the attention mechanism. The fused feature map $\boldsymbol{\mathit{Y}}_j$ can better capture both spatial information and spectral information. }
    \label{fig:attention}
\end{figure}


\subsection{Sparse Deformable Sequencing (SDS)}

The SDS approach is proposed to address the following two key issues. 

\begin{itemize}

\item \textbf{Sparsity:} Instead of using all tokens as conducted in classical scanning approaches as in Figure \ref{fig:sequence} (a)-(h), how to reduce the potential redundancy and computational cost by building sparse and compact sequence only using the most relevant tokens? 

\item \textbf{Learnability and Adaptability:} Instead of defining the sequence in deterministic predefined manners as in Figure \ref{fig:sequence} (a)-(h), how to adaptively build a deformable sequence in a learnable manner to overcome the rigidity of classical scanning approaches?

\end{itemize}

First, as illustrated in Figure \ref{fig:overview}, the proposed SDMamab is sparse, because the input sequences to the Mamba Blocks, i.e.,  $\overline{\boldsymbol{\mathit{Z}}}_j (5 \ tokens)$ and $\overline{\boldsymbol{\mathit{A}}}_j (3 \ tokens)$ have much less number of tokens than  $\boldsymbol{\mathit{Z}}_j (HW \ tokens)$ and $\boldsymbol{\mathit{A}}_j (C \ tokens)$ respectively. 

Second, in SDMamba, the order of tokens in $\overline{\boldsymbol{\mathit{Z}}}_j$ and $\overline{\boldsymbol{\mathit{A}}}_j$ are deformable and learnable, because two adaptive attention matrices, i.e., $SparseSpatialAttn$ and $SparseSpectralAttn$, are used to sort the tokens and identify a limited number of tokens. 

Therefore, the proposed SDMamba approach has sparse and deformable sequence patterns that can reduce redundancy, rigidity, and computational cost in classical Mamba models.

\subsection{Sparse Deformable Spatial Mamba Module (SDSpaM)}

Based on the proposed SDS approach, we design a SDSpaM module to focus on learning the spatial information in HSI. 

In Figure \ref{fig:overview}, the SDSpaM module takes as input $\mathit{Z_j}$, which has a total of $H*W$ tokens, with each token being a $C \times 1$ vector. Instead of using $\mathit{Z_j} \ (H*W \ tokens)$ directly as input to the $MambaBlock$, we generate a \textbf{Sparse Deformable Spatial Sequence}, denoted by 
$\overline{\boldsymbol{\mathit{Z}}}_j \ (5 \ tokens)$ to feed the $MambaBlock$, to reduce the potential redundancy and computational cost. $SparseSpatialAttn$, a $H \times W$ matrix, is used to identify these 5 tokens by sorting these tokens according to their relevance. So, the sequence is deformable and learnable, because $SparseSpatialAttn$ is adaptively learned from the data. The output of $MambaBlock$, denoted by $\underline{\boldsymbol{\mathit{Z}}}_j$, is scattered into the spatial dimensions of $\mathit{Z_j}$, which serves as a residual skip connection that is commonly used in neural network architectures.

More specifically, a stem layer is used to get the initial feature:
\begin{align}
    \mathbf{\mathit{Y_j}} &= \mathrm{GELU}\left(\mathrm{BatchNorm}\left(\text{Conv}(\mathbf{\mathit{X_j})}\right)\right) 
\end{align}

\noindent where $\mathbf{\mathit{X_j}}$ is input data cube and $\mathbf{\mathit{Y_j}}$ is the feature map.

We reshape $\mathbf{\mathit{Y_j}}$ into $\mathbf{\mathit{Z_j}}$, a $HW \times C$ matrix, with $HW$ being the number of tokens and each token being a $C \times 1$ vector. 

We then select the central token \(\mathbf{\mathit{z_j}} \in \mathbb{R}^{1 \times C}\) as an anchor to measure the cosine similarity with all tokens in $\mathbf{\mathit{Z_j}}$. 

The $i$th element in the sparse spatial attention matrix \(SparseSpatialAttn \in \mathbb{R}^{H \times W}\) can be calculated by the the following equation:
\begin{align}
    {\mathrm{SparseSpatialAttn}}_i &= \arccos \left( \frac{z_{j}^{\top} z_{i}}{\| z_{j}\| \| z_{i}\|} \right) 
\end{align}

\noindent where $\mathbf{\mathit{z_i}}$ is the $i$th token in $\mathbf{\mathit{Z_j}}$. 

We sort all elements in $SparseSpatialAttn$ according to their magnitudes and identify a small sorted subset of tokens to achieve sparse deformable token sequence. A sparsity ratio \(\lambda\) is used to control the size of the subset. Here we set \(\lambda = 0.3\). 


\subsection{Sparse Deformable Spectral Mamba Module (SDSpeM)}

Based on the proposed SDS approach, we design a SDSpeM module to focus on learning the spectral information in HSI. 

In Figure \ref{fig:overview}, the SDSpeM module also takes as input $\mathit{A_j}$, which has a total of $C$ tokens, with each token being a $H*W \times 1$ vector. Instead of using $\mathit{A_j} \ (C \ tokens)$ directly as input to the $MambaBlock$, we generate a \textbf{Sparse Deformable Spectral Sequence}, denoted by 
$\overline{\boldsymbol{\mathit{A}}}_j \ (3 \ tokens)$ to feed the $MambaBlock$, to reduce the potential redundancy and computational cost. 

$SparseSpectralAttn$, a $C \times 1$ vector, is used to sort these tokens and identify these 3 tokens. So, the sequence is deformable and learnable, because $SparseSpectralAttn$ is adaptively learned from the data. 

The output of $MambaBlock$, denoted by $\underline{\boldsymbol{\mathit{A}}}_j$, is scattered into the spatial dimensions of $\mathit{A_j}$, which serves as a residual skip connection that is commonly used in neural network architectures. 

More specifically, we reshape $\mathbf{\mathit{Y_j}}$ into $\mathbf{\mathit{A_j}}$, a $C \times HW$ matrix, with $C$ being the number of tokens and each token being a $HW \times 1$ vector. 

We then select a random token \(\mathbf{\mathit{a_j}} \in \mathbb{R}^{HW \times 1}\) as an anchor to measure the cosine similarity with all tokens in $\mathbf{\mathit{A_j}}$. 

The $i$th element in the sparse spatial attention matrix \(SparseSpectralAttn \in \mathbb{R}^{C \times 1}\) can be calculated by the the following equation:

\begin{align}
    {\mathrm{SparseSpectralAttn}}_i &= \arccos \left( \frac{a_{j}^{\top} a_{i}}{\| a_{j}\| \| a_{i}\|} \right) 
\end{align}

\noindent where $\mathbf{\mathit{a_i}}$ is the $i$th token in $\mathbf{\mathit{A_j}}$. 

We sort all elements in $SparseSpectralAttn$ according to their magnitudes and identify a small sorted subset of tokens to achieve sparse deformable token sequence. A sparsity ratio \(\lambda\) is used to control the size of the subset. Here we set \(\lambda = 0.3\).

\subsection{Attention Data Fusion Module}
Based on the outputs ($H \times W \times 256$) of the SDSpaM module and the SDSpeM module, we design an attention fusion approach to leverage the attention mechanism for improving the fusion of spatial information and spectral information. As illustrated in Figure \ref{fig:attention}, first, we use the output of SDSpaM to calculate $Q$ of size $(H*W) \times 256$, and use the SDSpeM output to calculate $K$ and $K$ with the same size. An attention matrix is achieved by multiplying $Q$ with $K$, which is then used to update $V$ for achieving the fused feature map. 

\section{Results and Analysis}


\subsection{Implementation schema}

We compare the proposed method with various state-of-the-art approaches, i.e., SSRN \cite{8061020}, SS-ConvNeXt \cite{10144309}, MTGAN \cite{MTGAN}, SSFTT \cite{SSFTT}, SSTN \cite{9565208}, GSC-ViT \cite{10472541}, MambaHSI \cite{10604894}, 3DSS-Mamba \cite{10703171}, HyperMamba \cite{liu2024hypermamba}, on some benchmark datasets, i.e., Indian Pines (IP) and the Pavia University (PU). For IP, we use 10\% and 10\% samples for training and validating, the rest of the samples for testing. For PU, we use 3\% and 1\% for training and validating, the rest of samples for testing. We use overall accuracy (OA), averaged accuracy (AA) and the kappa coefficient for evaluating the performance of the methods. For our method, we use a patch-size of $13$ for IP and $19$ for PU, batch-size of $64$ for both datasets, learning rate of $0.0001$, $100$ epoches, $256$ for hidden dimensions, $30\%$ sparsity ratio for both datasets.  

\begin{itemize}
    \item Indian Pines dataset was collected by the
 AVIRIS sensor over Northwestern Indiana, USA. This data consists of 145 × 145 pixels with 220 spectral bands covering the wavelength range of 400–2500 nm. In the experiment, 24 water-absorption bands and noise bands were removed, and 200 bands were selected. There are 16 investigated categories in this studied scene.

 \item Pavia University dataset was acquired by the ROSIS sensor over Pavia University and its surroundings, Pavia, Italy. This dataset has 103 spectral bands ranging from 430 to 860nm. Its spatial resolution is 1.3 m, and its image size is 610 × 340. Nine land-cover categories are covered
 \end{itemize}

\subsection{Results}

Table \ref{ablation_study} shows the influence of sparsity ratios and the
floating-point operations (FLOPs). As we can see, our sparse approach not only has much less FLOPS with reduced computational cost, but also higher accuracy than the classical dense token approach (i.e., w/o Sparsity in the table). In fact, using 5\% of the tokens (i.e., Sparsity ratio 0.05) outperforms the classical Mamba approach on the Indian pines dataset, and achieves comparable performance on the Pavia dataset, demonstrating the significant benefits of reducing the redundancy in the Mamba sequence. 

\begin{figure}[]
\centering
\includegraphics[width=0.49\textwidth]{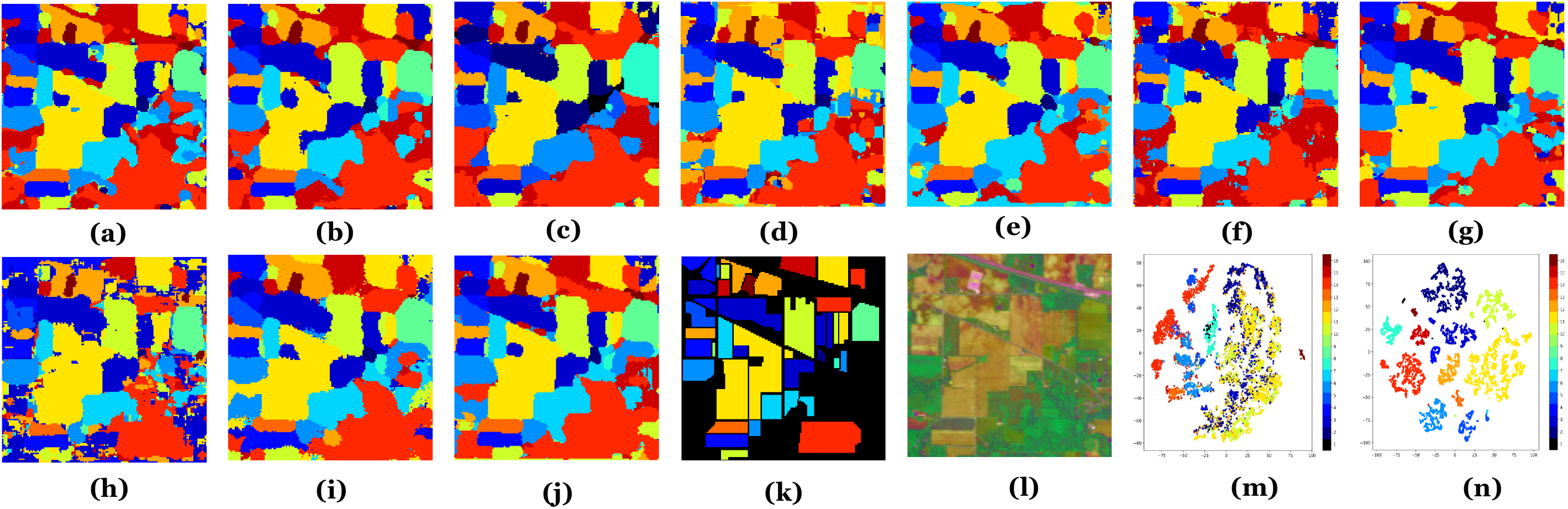}
\includegraphics[width=0.49\textwidth]{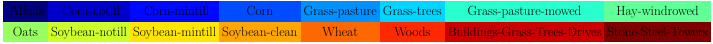}
\caption{The Indian Pines classification map generated by different methods. (a) SSRN (b) SS-ConvNeXt (c) MTGAN (d) SSFTT (e) SSTN (f) GSC-ViT (g) MammbaHSI (h) 3DSS-Mamba (i) HyperMamba (j) SDMamba (k) Ground Truth (l) RGB Image (m) TSNE features on labeled data (n) TSNE features on predicted pixel}
\label{fig:IP}
\end{figure}

Figure \ref{fig:IP} shows the classification maps achieved by different methods on the IP dataset. As we can see, the proposed approach achieves a map that is not only the most consistent with the classification map, but also better at delineating the boundaries and small classes. 

Table \ref{table:IP} shows the numerical results achieved by different methods on the the IP dataset. Our approach outperforms the other methods on all metrics. In particular, our approach achieves much better results on AA, indicating that the proposed approach outperforms the other approaches in terms of preserving and classifying the small classes.

\begin{figure}[]
\centering
\includegraphics[width=0.49\textwidth]{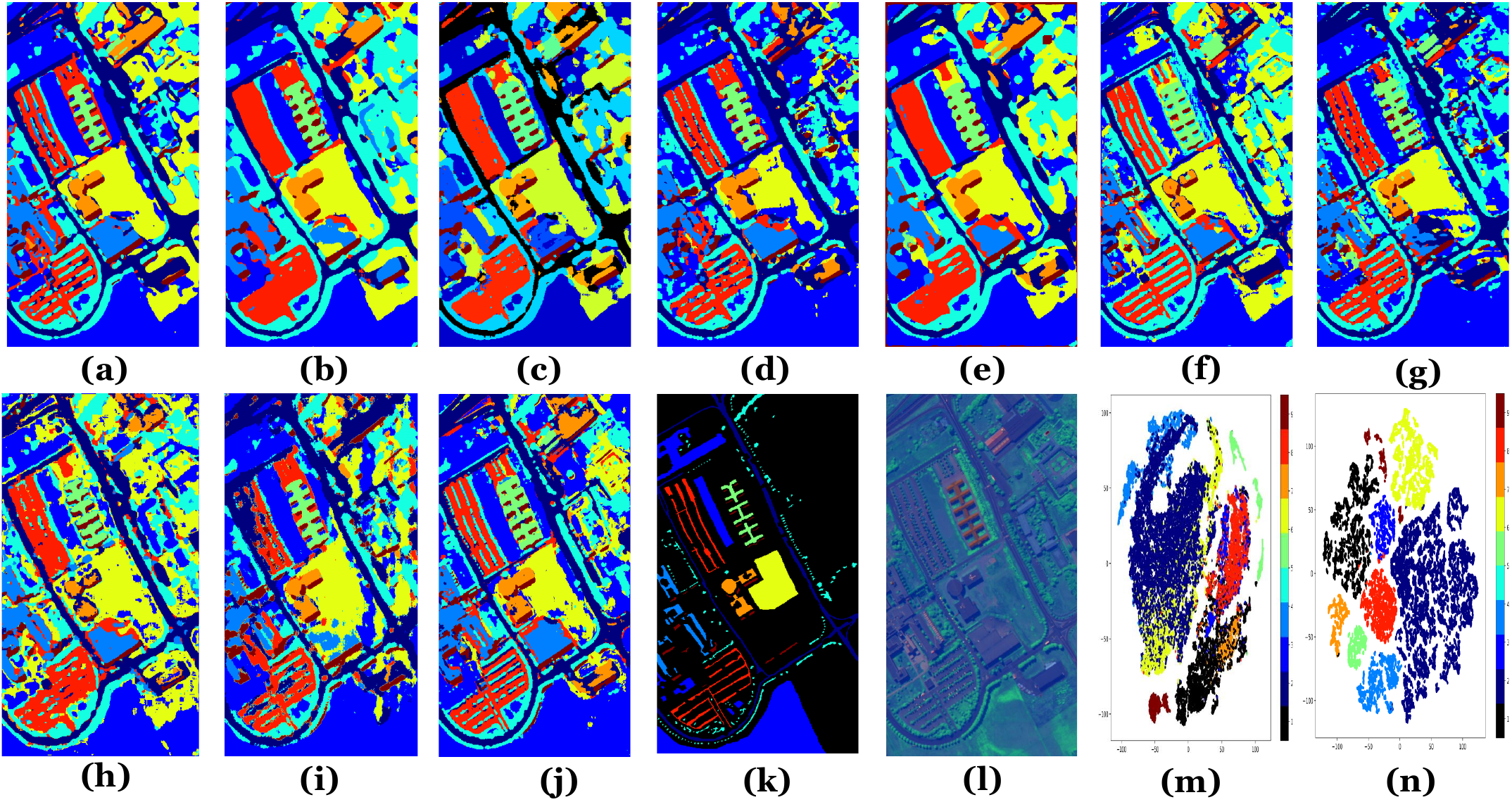}

\includegraphics[width=0.49\textwidth]{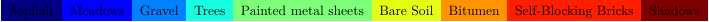}

\caption{The Pavia University classification map generated by different methods. (a) SSRN (b) SS-ConvNeXt (c) MTGAN (d) SSFTT (e) SSTN (f) GSC-ViT (g) MammbaHSI (h) 3DSS-Mamba (i) HyperMamba (j) SDMamba (k) Ground Truth (l) RGB Image (m) TSNE features on labeled data (n) TSNE features on predicted pixel}
\label{fig:PU}
\end{figure}

Figure \ref{fig:PU} shows the classification maps achieved by different methods on the PU dataset. As we can see, the proposed approach performs consistently well as the IP dataset. It achieves a map that is more consistent with the ground-truth map with better boundaries and small classes. 

Table \ref{table:PU} shows the numerical results achieved by different methods on the the PU dataset. As we can see, our approach consistently outperforms the other methods on all metrics. The fact that our approach achieves much better results on AA demonstrates that the proposed approach outperforms the other approaches in terms of preserving and classifying the small classes. 

Futrhermore, Figure \ref{fig:IP} (m)-(n), and Figure \ref{fig:PU} (m)-(n) show the TSNE visualization of features extracted by proposed SDMamba model. We can clearly see that our approach can disentangle the different classes that are hiding in the original space.

\begin{table*}
\caption{Quantitative performance of different classification methods in terms of OA, AA, $k$, as well as the accuracies for each class on the Indian Pines dataset with 10 \% training samples. The best results are in bold and colored shadow.}
\resizebox{\textwidth}{!}{
\begin{tabular}{ccc|cc|c|ccc|cccc}
\hline
\multicolumn{1}{c|}{\multirow{2}{*}{Class No.}} & \multicolumn{1}{c|}{\multirow{2}{*}{Train Number}} & \multirow{2}{*}{Test Number} & \multicolumn{2}{c|}{CNN-based} & GAN-based & \multicolumn{3}{c|}{Transformer-based} & \multicolumn{4}{c}{Mamba-based}                             \\ \cline{4-13} 
\multicolumn{1}{c|}{}                           & \multicolumn{1}{c|}{}                              &                              & SSRN      & SS-ConvNeXt        & MTGAN     & SSFTT  & SSTN         & GSC-ViT        & MambaHSI     & HyperMamba     & 3DSS-Mamba & \textbf{SD-Mamba}  \\ \hline
\multicolumn{1}{c|}{1}                          & \multicolumn{1}{c|}{5}                             & 37                           & 90.98     & 84.21              & 84.63     & 94.18  & 94.44        & 94.59          & 94.74        & \cellcolor[RGB]{251, 228, 213}\textbf{97.56} & 95.12      & 97.30           \\
\multicolumn{1}{c|}{2}                          & \multicolumn{1}{c|}{143}                           & 1142                         & 97.54     & \cellcolor[RGB]{251, 228, 213}\textbf{99.53}     & 97.20      & 94.9   & 97.99        & 95.19          & 98.08        & 98.44          & 93.22      & 99.82          \\
\multicolumn{1}{c|}{3}                          & \multicolumn{1}{c|}{83}                            & 664                          & 96.88     & 96.62              & 95.9      & 92.16  & 95.48        & 96.54          & 96.36        & 99.19          & 93.44      & \cellcolor[RGB]{251, 228, 213}\textbf{99.4}  \\
\multicolumn{1}{c|}{4}                          & \multicolumn{1}{c|}{24}                            & 189                          & 97.7      & 95.24              & 96.85     & 94.8   & 93.12        & \cellcolor[RGB]{251, 228, 213}\textbf{99.47} & 90.05        & 96.24          & 96.71      & \cellcolor[RGB]{251, 228, 213}\textbf{100.00}   \\
\multicolumn{1}{c|}{5}                          & \multicolumn{1}{c|}{48}                            & 387                          & 95.13     & 97.91              & 95.61     & 95.33  & 99.22        & 98.45          & 97.67        & \cellcolor[RGB]{251, 228, 213}\textbf{99.77} & 96.71      & 97.93          \\
\multicolumn{1}{c|}{6}                          & \multicolumn{1}{c|}{73}                            & 584                          & 99.25     & 99.85              & 98.48     & 96.66  & 99.66        & 98.29          & 99.49        & \cellcolor[RGB]{251, 228, 213}\textbf{99.85} & 95.40      & 99.32          \\
\multicolumn{1}{c|}{7}                          & \multicolumn{1}{c|}{3}                             & 22                           & 76.4      & \cellcolor[RGB]{251, 228, 213}\textbf{100.00}                & 12.00        & 84.92  & 68.18        & 77.27          & \cellcolor[RGB]{251, 228, 213}\textbf{100.00} & \cellcolor[RGB]{251, 228, 213}\textbf{100.00}   & 98.63      & \textbf{100.00}   \\
\multicolumn{1}{c|}{8}                          & \multicolumn{1}{c|}{48}                            & 382                          & 99.53     & 99.76              & 99.95     & 99.62  & \cellcolor[RGB]{251, 228, 213}\textbf{100.00}          & \cellcolor[RGB]{251, 228, 213}\textbf{100.00}            & \cellcolor[RGB]{251, 228, 213}\textbf{100.00} & \cellcolor[RGB]{251, 228, 213}\textbf{100.00}   & 96.97      & \cellcolor[RGB]{251, 228, 213}\textbf{100.00}   \\
\multicolumn{1}{c|}{9}                          & \multicolumn{1}{c|}{2}                             & 16                           & 55.29     & 92.86              & 61.76     & 77.83  & 78.57        & 93.75          & \cellcolor[RGB]{251, 228, 213}\textbf{100.00} & 82.35          & 72.22      & \cellcolor[RGB]{251, 228, 213}\textbf{100.00}   \\
\multicolumn{1}{c|}{10}                         & \multicolumn{1}{c|}{97}                            & 778                          & 96.19     & 97.81              & 95.43     & 92.21  & 98.97        & 97.17          & 97.04        & 98.4           & 92.80      & \cellcolor[RGB]{251, 228, 213}\textbf{99.74} \\
\multicolumn{1}{c|}{11}                         & \multicolumn{1}{c|}{245}                           & 1965                         & 98.29     & 88.78              & 98.44     & 97.52  & 97.91        & 99.75          & 99.29        & 99.46          & 96.92      & \cellcolor[RGB]{251, 228, 213}\textbf{99.13} \\
\multicolumn{1}{c|}{12}                         & \multicolumn{1}{c|}{59}                            & 475                          & 97.97     & 97.92              & 95.10     & 90.48  & \cellcolor[RGB]{251, 228, 213}\textbf{100.00} & 95.16          & 99.16        & 99.44          & 90.64      & 98.53          \\
\multicolumn{1}{c|}{13}                         & \multicolumn{1}{c|}{20}                            & 164                          & 99.68     & \cellcolor[RGB]{251, 228, 213}\textbf{100.00}       & 98.76     & 96.59  & 99.39        & \cellcolor[RGB]{251, 228, 213}\textbf{100.00}   & 99.39        & 100.00            & 95.14      & \cellcolor[RGB]{251, 228, 213}\textbf{100.00}   \\
\multicolumn{1}{c|}{14}                         & \multicolumn{1}{c|}{126}                           & 1012                         & 99.57     & 99.91              & 99.12     & 98.67  & 99.8         & 99.7           & 99.11        & 99.74          & 99.65      & \cellcolor[RGB]{251, 228, 213}\textbf{100.00}   \\
\multicolumn{1}{c|}{15}                         & \multicolumn{1}{c|}{39}                            & 308                          & 98.01     & 99.71              & 98.10     & 83.48  & 99.35        & 96.44          & \cellcolor[RGB]{251, 228, 213}\textbf{100.00} & 99.42          & 95.68      & \cellcolor[RGB]{251, 228, 213}\textbf{100.00}   \\
\multicolumn{1}{c|}{16}                         & \multicolumn{1}{c|}{9}                             & 75                           & 97.38     & 97.53              & 92.50     & 89.93  & 89.33        & 97.3           & 85.33        & 97.62          & 85.71      & \cellcolor[RGB]{251, 228, 213}\textbf{98.67} \\ \hline
\multicolumn{3}{c|}{OA(\%)}                                                                                                         & 97.81     & 96.29              & 97.20     & 95.12  & 98.23        & 97.94          & 98.28        & 99.18          & 95.53      & \cellcolor[RGB]{251, 228, 213}\textbf{99.44} \\
\multicolumn{3}{c|}{AA(\%)}                                                                                                         & 93.49     & 96.73              & 88.80     & 92.46  & 94.49        & 96.19          & 97.23        & 97.97          & 94.9       & \cellcolor[RGB]{251, 228, 213}\textbf{99.36} \\
\multicolumn{3}{c|}{Kappa(\%)}                                                                                                      & 97.74     & 95.96              & 97.08     & 94.76  & 97.98        & 97.65          & 98.21        & 99.06          & 93.42      & \cellcolor[RGB]{251, 228, 213}\textbf{99.36} \\ \hline
\end{tabular}
\label{table:IP}
}
\end{table*}

\begin{table*}[h]
\caption{Quantitative performance of different classification methods in terms of OA, AA, $k$, as well as the accuracies for each class on the Pavia University dataset with 3 \% training samples. The best results are in bold and colored shadow.}
\resizebox{1\textwidth}{!}{
\begin{tabular}{ccccc|c|ccc|cccc}
\hline
\multicolumn{1}{c|}{\multirow{2}{*}{Class No.}} & \multicolumn{1}{c|}{\multirow{2}{*}{Train Number}} & \multicolumn{1}{c|}{\multirow{2}{*}{Test Number}} & \multicolumn{2}{c|}{CNN-based}  & GAN-based      & \multicolumn{3}{c|}{Transformer-based}             & \multicolumn{4}{c}{Mamba-based}                                  \\ \cline{4-13} 
\multicolumn{1}{c|}{}                           & \multicolumn{1}{c|}{}                              & \multicolumn{1}{c|}{}                             & SSRN           & SS-ConvNeXt    & MTGAN          & SSFTT          & SSTN            & GSC-ViT         & MambaHSI        & HyperMamba      & 3DSS-Mamba & \textbf{Ours}   \\ \hline
\multicolumn{1}{c|}{1}                          & \multicolumn{1}{c|}{199}                           & \multicolumn{1}{c|}{6366}                         & 98.74          & 98.79          & 98.68          & \cellcolor[RGB]{251, 228, 213}\textbf{99.67} & 98.32           & 97.34           & 98.66           & 99.30 & 97.05      & 98.59           \\
\multicolumn{1}{c|}{2}                          & \multicolumn{1}{c|}{559}                           & \multicolumn{1}{c|}{17904}                        & 99.74          & 99.73 & 99.79          & 99.94          & 99.85           & 99.02           & 99.94           & \cellcolor[RGB]{251, 228, 213}\textbf{99.99}  & 98.02      & \cellcolor[RGB]{251, 228, 213}\textbf{99.99}  \\
\multicolumn{1}{c|}{3}                          & \multicolumn{1}{c|}{63}                            & \multicolumn{1}{c|}{2015}                         & 90.69          & 94.66          & 97.16          & 90.71          & \cellcolor[RGB]{251, 228, 213}\textbf{97.47}  & 84.54           & 94.99           & 88.90           & 98.04      & 94.94  \\
\multicolumn{1}{c|}{4}                          & \multicolumn{1}{c|}{92}                            & \multicolumn{1}{c|}{2941}                         & 97.30          & 96.21          & 97.44          & 92.56          & 93.54           & 96.67  & 95.68           & 94.62           & 88.26      & \cellcolor[RGB]{251, 228, 213}\textbf{97.65}  \\
\multicolumn{1}{c|}{5}                          & \multicolumn{1}{c|}{40}                            & \multicolumn{1}{c|}{1292}                         & 99.89          & 99.86          & 99.59          & 99.41          & \cellcolor[RGB]{251, 228, 213}\textbf{100.00} & \cellcolor[RGB]{251, 228, 213}\textbf{100.00} & \cellcolor[RGB]{251, 228, 213}\textbf{100.00} & \cellcolor[RGB]{251, 228, 213}\textbf{100.00} & 97.85      & \cellcolor[RGB]{251, 228, 213}\textbf{100.00} \\
\multicolumn{1}{c|}{6}                          & \multicolumn{1}{c|}{151}                           & \multicolumn{1}{c|}{4828}                         & 99.50          & 99.70          & \cellcolor[RGB]{251, 228, 213}\textbf{99.73} & 95.39          & 99.19           & 98.96           & 99.48           & 99.69  & 99.57      & 99.15           \\
\multicolumn{1}{c|}{7}                          & \multicolumn{1}{c|}{40}                            & \multicolumn{1}{c|}{1277}                         & 96.11          & 98.61          & 95.97          & \cellcolor[RGB]{251, 228, 213}\textbf{100.00}         & \cellcolor[RGB]{251, 228, 213}\textbf{100.00} & 99.12           & 99.61  & 96.05 & 75.89      & 97.10  \\
\multicolumn{1}{c|}{8}                          & \multicolumn{1}{c|}{110}                           & \multicolumn{1}{c|}{3534}                         & 97.02          & 97.41          & 98.83          & 93.86          & 96.72           & 99.08           & 98.64  & 82.08  & 98.12      & \cellcolor[RGB]{251, 228, 213}\textbf{99.92}  \\
\multicolumn{1}{c|}{9}                          & \multicolumn{1}{c|}{28}                            & \multicolumn{1}{c|}{909}                          & \cellcolor[RGB]{251, 228, 213}\textbf{98.97} & 97.23          & 96.61          & 96.09          & 86.70           & 98.76           & \cellcolor[RGB]{251, 228, 213}\textbf{98.79}  & 98.48           & 91.62      & 98.68  \\ \hline
\multicolumn{3}{c}{OA(\%)}                                                                                                                               & 98.58          & 98.79          & 99.03          & 97.76          & 98.41           & 97.91           & 99.00           & 97.22           & 96.52      & \cellcolor[RGB]{251, 228, 213}\textbf{99.14}  \\
\multicolumn{3}{c}{AA(\%)}                                                                                                                               & 97.55          & 98.02          & 98.20          & 96.40          & 96.87           & 97.05           & 98.42           & 95.46           & 93.82      & \cellcolor[RGB]{251, 228, 213}\textbf{98.45}  \\
\multicolumn{3}{c}{Kappa(\%)}                                                                                                                            & 98.16          & 98.79          & 98.76          & 97.10          & 97.90  & 97.23           & 98.11           & 96.31           & 95.40      & \cellcolor[RGB]{251, 228, 213}\textbf{98.85}  \\ \hline
\end{tabular}
}
\label{table:PU}
\end{table*}

\begin{table}[]
\caption{Ablation study for different sparsity ratio and FLOPs}
\resizebox{0.5\textwidth}{!}{
\begin{tabular}{ccccccc|cc}
\hline
                                 & \multicolumn{3}{c}{Indian Pines}             & \multicolumn{3}{c|}{Pavia University}                          & \multicolumn{2}{c}{}                                                         \\ \cline{2-7}
\multirow{-2}{*}{Sparsity ratio} & OA    & AA    & {Kappa} & OA             & AA             & {Kappa} & \multicolumn{2}{c}{\multirow{-2}{*}{FLOPS}}                                  \\ \hline
0.05                             & 98.95 & 98.46 & {98.80}  & 98.64          & 97.25          & 98.2                &                                       &                                      \\
0.1                              & 99.1  & 98.74 & {98.97} & 99.31 & 98.70           & 99.08                        & \multirow{-2}{*}{with Sparsity ratio} & \multirow{-2}{*}{w/o Sparsity ratio} \\
0.3                              & 99.44 & 99.36 & 99.36                        & 99.14          & 98.45          & 98.85                        & 172.41M                      & 416.23M                              \\
0.5                              & 99.32 & 99.49 & 99.22                        & 98.76          & 98.06          & 98.35                        & -                                     & -                           \\
0.7                              & 99.2  & 99.36 & 99.08                        & 99.02          & 98.61          & 98.7                         & -                            & -                           \\ \hline
w/o Sparsity ratio               & 98.82 & 97.96 & 98.65                        & 98.80          & 97.91 & 98.41                        & -                                     & -        \\ \hline                            
\end{tabular}
}
\label{ablation_study}
\end{table}

\section{Conclusion}
In this paper, we have presented a Sparse Deformable Mamba (SDMamba) approach to enhance the HSI classification. We have the following contributions. First, an efficient Sparse Deformable Sequencing (SDS) approach has been designed to learn the "optimal" sequencing, which not only optimize the learning capacity of the Mamba model but also increases its efficiency with less computations. Second, the SDS approach was integrated with the spatial module and the spectral module, leading to two dedicated HSI feature learning modules, i.e., the Sparse Deformable Spatial Mamba Module (SDSpaM) and Sparse Deformable Spectral Mamba Module (SDSpeM), which are dedicated to learning the spatial context information and SDspeM focusing on the spectral information respectively. Last, a novel feature fusion approach was designed based on the attention mechanism which can efficiently integrate the output of the SDSpaM and SDSpeM for HSI classification. The proposed approach was tested on the Indian Pines and Pavia University HSI  datasets in comparison with various other state-of-the-art approaches, demonstrating that our approach outperformed others in terms of both accuracy and computational cost. 


\bibliographystyle{IEEEtran}
\bibliography{IEEEabrv,references}

\end{document}